\documentclass{article}
\usepackage{spconf,amsmath,graphicx,multirow,bbding,booktabs,amssymb,bm,bbm,hyperref}
\usepackage[american]{babel}

\hypersetup{
    colorlinks=true,
    linkcolor=blue,
    filecolor=blue,      
    urlcolor=blue,
    citecolor=cyan,
}

\title{Video Individual Counting With Implicit One-to-Many Matching}
\name{Xuhui Zhu$^\dagger$, Jing Xu$^\ddagger$, Bingjie Wang$^\S$, Huikang Dai$^\ddagger$, Hao Lu$^\dagger$\thanks{H. Lu is the corresponding author. Part of work was done when B. Wang was with Huazhong University of Science and Technology.}}
\address{$^\dagger$National Key Laboratory of Multispectral Information Intelligent Processing Technology\\
School of AIA, Huazhong University of Science and Technology, China\\
$^\ddagger$FiberHome Telecommunication Technologies Co., Ltd., China\\
$^\S$Department of Computer Science, University of Rochester, Rochester, USA\\
{\small\textsf{\{xuhuizhu,hlu\}@hust.edu.cn}} 
}

\begin{document}
%
\maketitle
\begin{abstract}
Video Individual Counting (VIC) is a recently introduced task that aims to estimate pedestrian flux from a video. It extends conventional Video Crowd Counting (VCC) beyond the per-frame pedestrian count. In contrast to VCC that only learns to count repeated pedestrian patterns across frames, the key problem of VIC is how to identify co-existent pedestrians between frames, which turns out to be a correspondence problem. Existing VIC approaches, however, mainly follow a one-to-one (O2O) matching strategy where the same pedestrian must be exactly matched between frames, leading to sensitivity to appearance variations or missing detections. In this work, we show that the O2O matching could be relaxed to a one-to-many (O2M) matching problem, which better fits the problem nature of VIC and can leverage the social grouping behavior of walking pedestrians. We therefore introduce OMAN, a simple but effective VIC model with implicit One-to-Many mAtchiNg, featuring an implicit context generator and a one-to-many pairwise matcher. Experiments on the SenseCrowd and CroHD benchmarks show that OMAN achieves the state-of-the-art performance. Code is available at \href{https://github.com/tiny-smart/OMAN}{OMAN}.
\end{abstract}
\section{Introduction}

\label{sec:intro}
\begin{figure}[!t]
  \centering
  \centerline{\includegraphics[width=8.5cm]{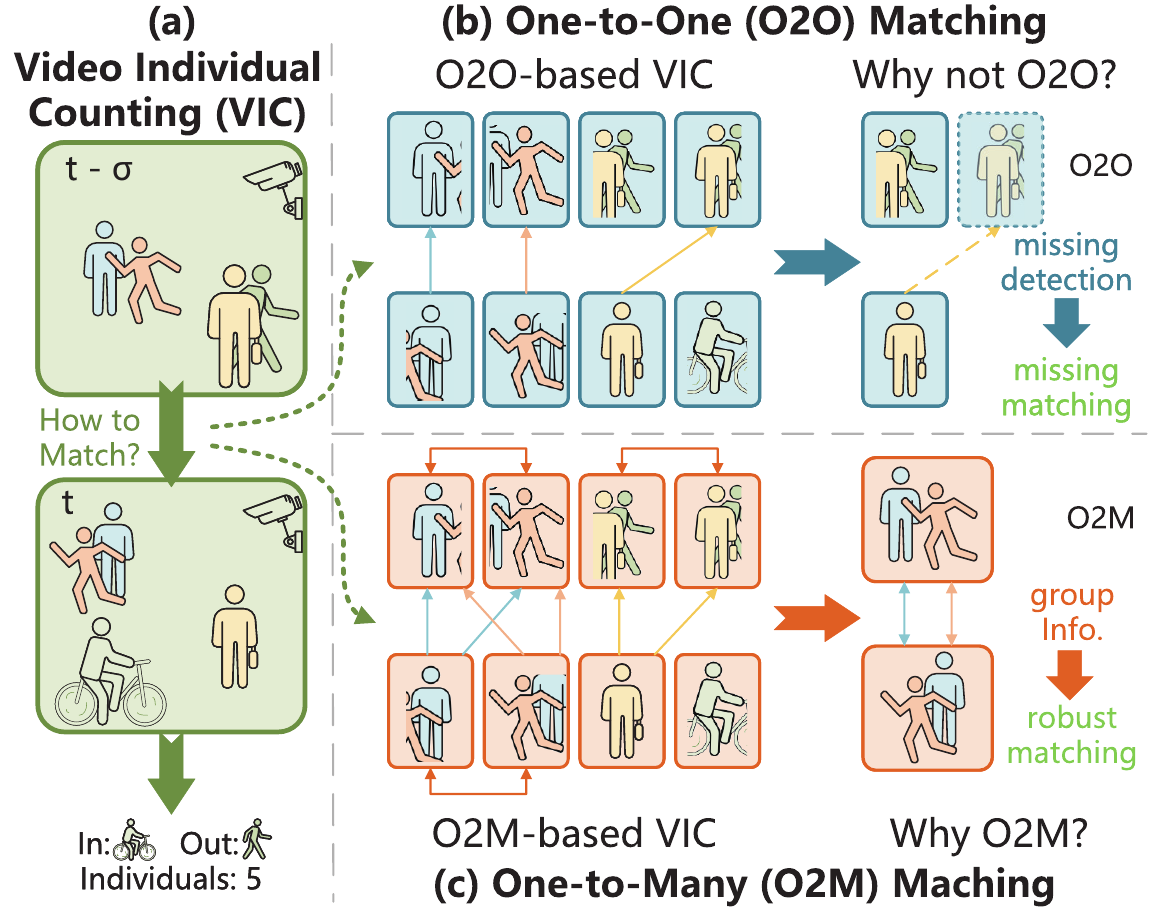}}
  \vspace{-10pt}
  \caption{ 
  \textbf{One-to-one (O2O) matching versus one-to-many (O2M) matching for video individual counting (VIC)}.
  (a) Addressing VIC needs to identify co-existent pedestrians between frames,
  while (b) existing VIC methods typically applies an O2O matching strategy, which can suffer from missing detections due to occlusions. 
  (c) Inspired by the grouping behavior of walking pedestrians, we relax the O2O matching to O2M matching such that an individual can assign to a pedestrian group to enhance matching robustness. 
  }
\label{fig:VIC}
\vspace{-10pt}
\end{figure}

Modern urbanization and population growth have led to increasingly crowded public spaces, raising concerns about public security. To reduce risks, video surveillance and vision-based techniques such as Video Crowd Counting (VCC)~\cite{LSTN,FTAN,PET} have been widely adopted to monitor the crowd. However, VCC can only address per-frame pedestrian counting, but cannot infer pedestrian flux within a period of time. To address this, cross-line crowd counting, a.k.a. Line of Interest (LOI)~\cite{LOI-2013,LOI-2016,LOI-2019}, and Multiple Object Tracking (MOT)~\cite{HeadHunter-T,FairMOT,PHDTT,ByteTrack} are introduced or repurposed to acquire such information. Yet, LOI cannot capture pedestrians in different directions and the computational cost of MOT can increase with the number of pedestrians detected. 

\begin{figure}[!t]
  \centering
  \centerline{\includegraphics[width=8.5cm]{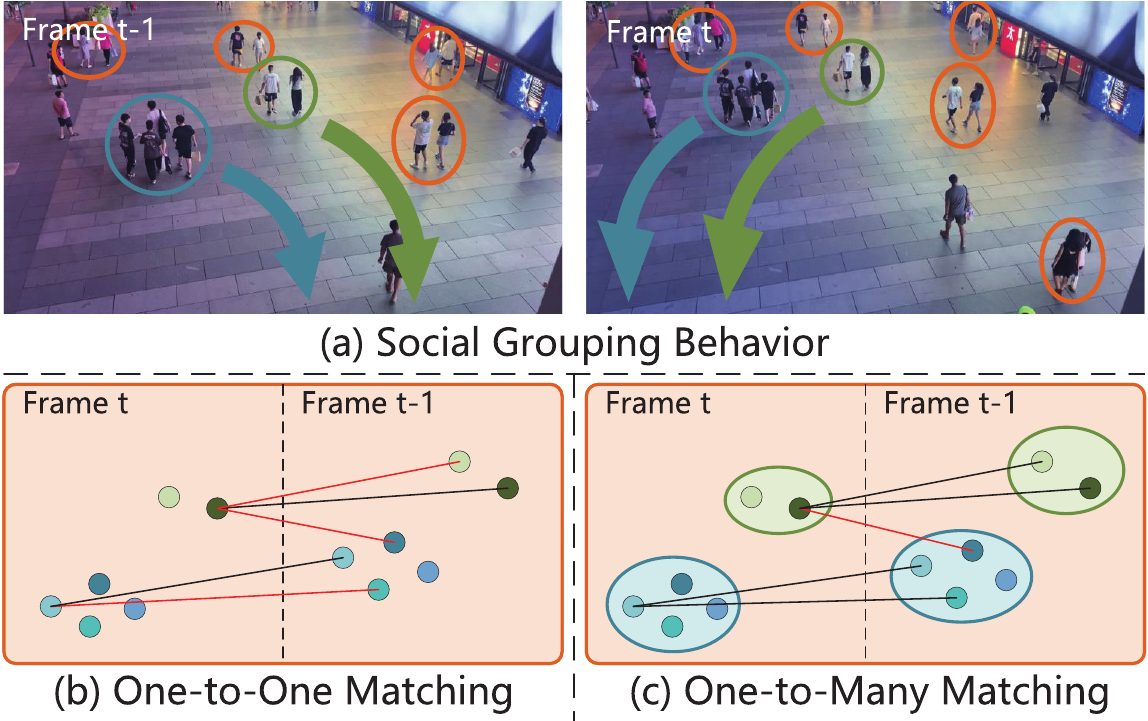}}
  \vspace{-10pt}
  \caption{
  \textbf{Motivation behind one-to-many matching}. 
  (a) Pedestrians tend to walk in groups. However, (b) one-to-one (O2O) matching fails to utilize this rule, leading to many mismatches. (c) We formulate pedestrian matching as a one-to-many problem, which exploits the social grouping behavior of pedestrians.}
\label{fig:match}
\vspace{-10pt}
\end{figure}

To better estimate pedestrian flux beyond LOI and MOT, a new task termed Video Individual Counting (VIC) is introduced recently~\cite{DRNet}, as shown in Fig.~\ref{fig:VIC}(a). DRNet~\cite{DRNet}, being one of the first baselines of this task, formulates VIC as a problem of pedestrian matching and predicts both pedestrian inflows and pedestrian outflows. While it demonstrates superior performance over MOT approaches, it still follows the problem setting of MOT and needs the identity labels of individuals. CGNet~\cite{CGNet}, with a deeper insight that VIC is about matching rather than tracking, replaces the identity annotations with category ones and proposes a group-level matching loss for supervision. To execute matching, both DRNet~\cite{DRNet} and CGNet~\cite{CGNet} adopt the idea of explicit O2O bipartite matching to identify co-existent pedestrians between adjacent frames, such as Hungarian matching~\cite{Hungarian} and optimal transport~\cite{OT}. Alternatively, two recent approaches, FMDC~\cite{FMDC} and PDTR~\cite{PDTR}, exploit regression to 
match the same pedestrian between frames, but their core idea, especially the training strategy, still follows O2O matching. O2O matching, however, highly depends on discriminative individual feature representations, leading to sensitivity to appearance variations due to occlusion, motion, and pose (Fig.~\ref{fig:VIC}(b)). In particular, an empirical threshold is often required to execute bipartite matching, further increasing the parameter sensitivity. To address this, we argue that \emph{group context} should be explored. As shown in Fig.~\ref{fig:match}(a), pedestrians often exhibit a social grouping behavior, that is, pedestrians tend to walk in a group. Even if not walking with a partner, 
pedestrians also share similar velocity and direction with their nearby neighbors. It seems more reasonable to model pedestrian groups instead of individuals to address VIC.

In this work, we relax the O2O matching into an O2M matching problem to harness the group context. In contrast to O2O matching that penalizes all wrong matches (Fig.~\ref{fig:VIC}(b)), O2M matching enables individual-to-group assignment (Fig.~\ref{fig:VIC}(c)), which can leverage shared appearance and contexts between pedestrians, particularly when occlusions occur. Indeed \textit{the nature of the VIC problem tells us that it is not necessary to match exactly the same pedestrian}. Technically, 
our idea is to implement O2M matching in a coarse-to-fine manner, 
first encoding pedestrians features into descriptors, then 
informing pedestrian features with contextual information at the feature level, and finally confirming them at the group level.
Inspired by a recent observation that self-attention can be decoupled to execute implicit similarity matching~\cite{SCAPE,CACViT}, we first use standard transformer encoder blocks to search for potential pedestrian matches at the feature level.
Finally, for 
all pairwise potential matches, we resort to a simple Multi-Layer Perceptron (MLP) to 
achieve the O2M assignment.

By encapsulating the O2M matching idea above into two network modules, featured by an Implicit Context Generator (ICG) and a One-to-Many Pairwise Matcher (OMPM), we introduce a simple yet effective solution termed One-to-Many mAtchiNg (OMAN) to VIC. Experiments on the SenseCrowd~\cite{SenseCrowd} and CroHD~\cite{HeadHunter-T} benchmarks reveal that OMAN outperforms the state-of-the-art approach, respectively, reducing MAE, MSE, and WRAE by $13.5\%$, $12.5\%$, and $2.6\%$ on the SenseCrowd dataset and by $28.8\%$, $18.2\%$ and $22.4\%$ on the CroHD dataset. Ablation studies are also conducted to justify the contribution of each module. Our results suggest that the \textit{O2M matching is a better matching paradigm than O2O to address VIC}, and OMAN indicates yet another strong baseline for VIC. 

\section{Proposed Method}
\label{sec:prop}
\begin{figure*}[!t]
  \centering
  \centerline{\includegraphics[width=16.5cm]{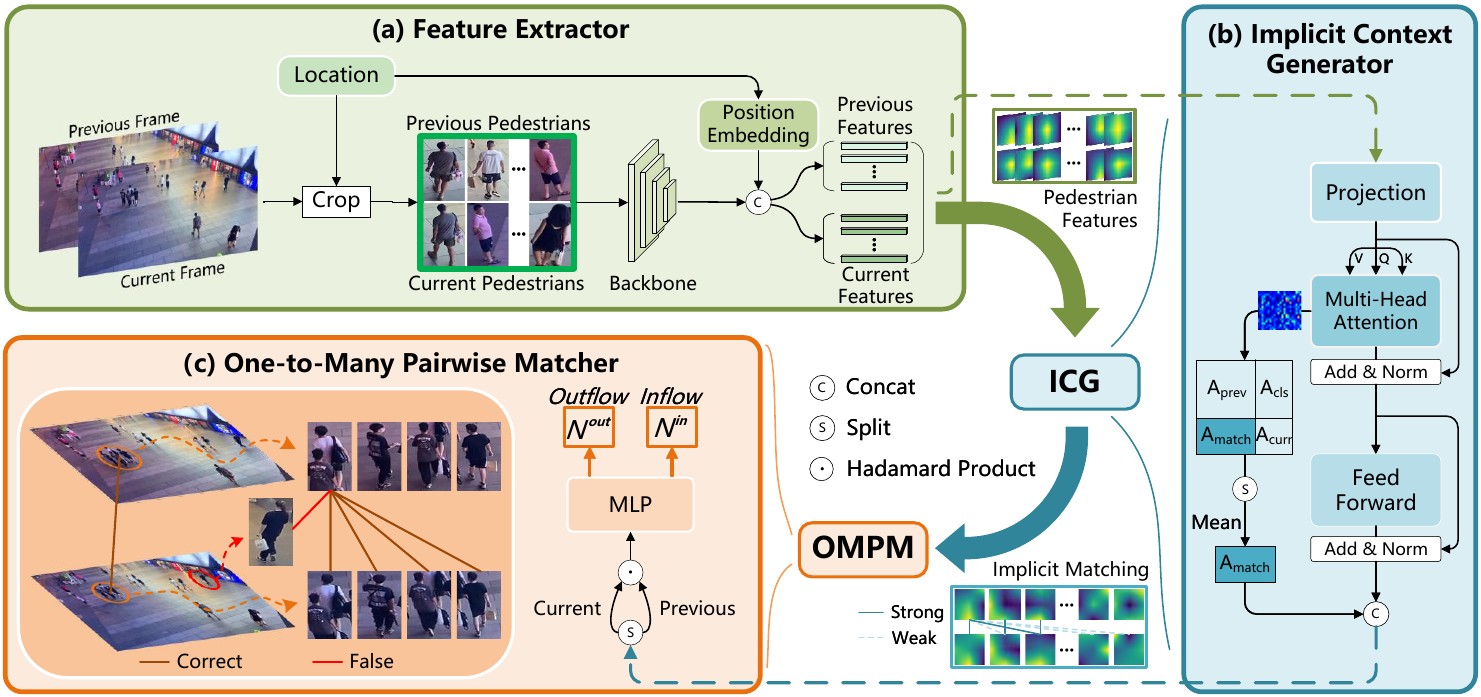}}
  \vspace{-5pt}
  \caption{\textbf{Technical pipeline of OMAN}. After (a) individuals are located and cropped with pre-provided location, their features are extracted with position embedding. We then apply (b) ICG to execute implicit context enhancement with decoupled self-attention maps and resort to (c) OMPM to implement one-to-many assignment, 
  generating individual-to-group matches and yielding the inflow and outflow numbers.}

\label{fig:Pipeline}
\end{figure*}

The technical pipeline of OMAN is shown in Fig.~\ref{fig:Pipeline}. 
Given the location of each pedestrian predicted by a pre-trained locator, following CGNet~\cite{CGNet}, we first crop the pedestrian patches from the frames. The pedestrian features are then extracted with position embedding, after which an ICG module with multi-head decoupled self-attention is used to implicitly provide the group context to inform potential pedestrian matches, and finally an OMPM module 
is used for individual-to-group identification. 
In what follows, we first define the problem setting and then explain the technical details.

\subsection{Problem Formulation}
\label{ssec:prop1}
Given a video sequence $I=\left \{ I_{0}, I_{1}, ..., I_{T} \right \}$ with $T+1$ frames, where the $t$-th frame $I_{t}$ contains $N_{t}$ pedestrians, the video is sampled with an interval of $\sigma$ seconds for efficiency. By coupling adjacent frames as pairs $\left \{I_{(k-1)\sigma}, I_{k\sigma} \right \}_{k=1}^{T/\sigma}$, in the $k$-th frame pair, \textit{pedestrians} who do not exist in $ I_{(k-1)\sigma}$, but show up in $I_{k\sigma}$ for the first time, are denoted as \textit{inflows}, with the inflow number $ N_{k\sigma}^{\tt in}$ counted. In contrast, pedestrians visible in $I_{(k-1)\sigma}$ but disappearing in $ I_{k\sigma}$ are regarded as \textit{outflows}, with the outflow number $N_{(k-1)\sigma}^{\tt out}$ also counted. 
The goal of VIC is to count unique pedestrians from a video sequence, which can be formulated by
\begin{equation}
    \label{eq:VIC}
    N_{\tt total} = N_{0} + \sum_{k=1}^{\tt T/\sigma}  N_{k\sigma}^{\tt in},
\end{equation}
where $N_{\tt total}$ is the total number of unique pedestrians, and $N_{0}$ is the first-frame pedestrian count. 

\subsection{Implicit Context Generator}

\label{ssec:prop2}

Inspired by the recent observations~\cite{CACViT,SCAPE} that self-attention can be decoupled to capture the implicit token relation for similarity matching, we introduce an ICG module to generate the group context and build the pedestrian relation between frames to inform potential pedestrian matches.

Specifically, the feature map $\mathcal F_{t}$ of the $t$-th frame with position embedding in Fig.~\ref{fig:Pipeline}(b) is first projected to $\bm F_{t}$. 
By concatenating all pedestrian features of the $t$-th and $(t-1)$-th frames, denoted by $\bm F_{t-1,t}$, $\bm F_{t-1,t}$ is then fed to the self-attention block and is normalized and projected to the query $Q$, key $K$ and value $V$. 
After self-attention, a certain attention map, indicating the affinity between tokens, can be split into four sub-attention maps~\cite{CACViT}, that is, $\bm A_{\tt prev}$,  $\bm A_{\tt cls}$,  $\bm A_{\tt match}$, and $\bm A_{\tt curr}$, as shown in the Fig.~\ref{fig:Pipeline}(b). Among them, $\bm A_{\tt prev}$ and $\bm A_{\tt curr}$ respectively indicate the pedestrian similarity in the previous frame (here is $\bm F_{t-1}$) and in the current frame (here is $\bm F_{t}$), while $\bm A_{\tt match}$ and $\bm A_{\tt cls}$ both represent the correlation between pedestrians in two frames. To characterize the group context, we average all $\bm A_{\tt match}$'s to $\bar{\bm A}_{\tt match}$. By concatenating $\bar{\bm A}_{\tt match}$ with the output of multi-head self-attention, 
we have the context-informed representation
\begin{equation}
    \label{eq:SAFR}
\bm F_{t-1,t}^{^{\prime}} = {\tt attn}(\bm F_{t-1,t})\oplus \bar{\bm A}_{\tt match}\,,
\end{equation}
where $\oplus$ denotes the concatenation operator,  and ${\tt attn}(\cdot )$ is multi-head self-attention function. 

According to the output of Fig.~\ref{fig:Pipeline}(b), 
$\bm F_{t-1,t}^{^{\prime}}$ informed by $\bar{A}_{\tt match}$ indicates the degree of connection of pedestrians between frames, which facilitates subsequent O2M matching.

\subsection{One-to-Many Pairwise Matcher}
\label{ssec:prop3}

After informing pedestrians about the group context, we turn to an OMPM module to implement the O2M assignment. 
By assigning an individual to a group, one can harness the social grouping behavior of pedestrians, especially in crowded scenes. 
This benefit is also mentioned in some related tasks such as ReID~\cite{Group-reid} and MOT~\cite{TransMOT,SCGTracker}. 

We formulate the O2M assignment as a classification problem. Formally, let $m=N_{t-1}$ and $n=N_{t}$ denote the pedestrian count in the $(t-1)$-th and the $t$-th frame, respectively. 
We first split the context-enhanced cross-frame pedestrian representation $\bm F_{t-1,t}^{^{\prime}}$ into single-frame representations 
$\bm F_{t-1}^{^{\prime}}=\{\bm f_{t-1}^{1}, \bm f_{t-1}^{2}, ..., \bm f_{t-1}^{m}\}$ and $\bm F_{t}^{^{\prime}}=\{\bm f_{t}^{1}, \bm f_{t}^{2}, ..., \bm f_{t}^{n}\}$. Then we compute the similarity vector 
between $\bm f_{t-1}^{i}$ and $\bm f_{t}^{j}$ by $\bm f_{t-1}^{i}\odot \bm f_{t}^{j}$, 
where $i=1,...,m$, $j=1,...,n$, and $\odot$ is the Hadamard product. Finally, we feed $\bm f_{t-1}^{i}\odot \bm f_{t}^{j}$ into an MLP to 
generate the probability of a successful match $p_{t-1,t}^{ij}$ between the two using a $\tt sigmoid$ function by 
\begin{equation}
    \label{eq:match}
    p_{t-1,t}^{ij} = {\tt sigmoid}({\tt MLP}(\bm f_{t-1}^{i}\odot \bm f_{t}^{j}))\,.
\end{equation}
In this way, the inflow count $N_t^{\tt in}$ of the $t$-th frame and the outflow count $N_{t-1}^{\tt out}$ of the $(t-1)$-th frame can be inferred by
\begin{align}
    \label{eq:match}
    N_t^{\tt in}&=N_{t}-\sum_{i,j}\lfloor p_{t-1,t}^{ij} \rceil\,,\\
    N_{t-1}^{\tt out}&=N_{t-1}-\sum_{i,j}\lfloor p_{t-1,t}^{ij}\rceil\,,
\end{align}
where $\lfloor \cdot \rceil$ is the rounding operator that rounds a number to the nearest integer.

During inference, a pedestrian in the $t$-th frame is allowed to match a different one in the $(t-1)$-th frame, as long as they are 
treated as a group, as shown in Fig.~\ref{fig:match}(c). 
Conceptually, the differences between the constraints for O2O and O2M matching are 
\begin{equation}
    \label{eq:O2O}
    \begin{aligned}
    {\tt O2O:}~~~& {\bm M}\mathbbm{1}_{m}\leq 1,\quad{\bm M}^T\mathbbm{1}_{n}\leq 1\\
    {\tt O2M:}~~~& {\bm M}\mathbbm{1}_{m}\leq k
    \end{aligned}\,,
\end{equation}
where $\bm{M}_{n\times m}$ represents matching matrix, $\mathbbm{1}_{n}$ is an $n$-dimensional column vector of ones, and $k$ is the maximum size of a group.

\subsection{Loss Function}
\label{ssec:prop4}
Following CGNet~\cite{CGNet}, we employ the Optimal Transport (OT) loss $\mathcal{L}_{\tt OT}$ for group-level contrastive learning under weak supervision.
In addition, all pairwise matching pairs 
are sent to the MLP, and the cross entropy loss $\mathcal{L}_{\tt cls}$ is used to supervise the training of OMPM. Here, for a pedestrian in a frame, we group this pedestrian and the neighboring ones within a certain normalized distance, $0.2$ for example, in adjacent frames. Then all pairwise matches within this group are considered positive matches, while all remaining ones are negative samples. Furthermore, 
we also add a Kullback-Leibler divergence loss $\mathcal{L}_{\tt KL}$ to impose a global constraint on the similarity distribution between predicted and ground-truth matching pairs, which empirically improves both the training stability and the performance. Mathematically, by discretizing the similarity values of the predicted matches and ground-truth matches into $K$ bins, with the probabilities $P(k)$ and $Q(k)$ representing the normalized frequency of similarity values that fall within each bin for predicted and ground-truth similarities, respectively. Then $\mathcal{L}_{\tt KL}$ is defined by  
\begin{equation}
    \label{eq:KL}
    \mathcal{L}_{\tt KL} = \sum_{k=1}^KP(k)\log\frac{P(k)}{Q(k)+\epsilon}\,,
\end{equation}
where \( \epsilon \) is a smoothing term to avoid zero division. Finally, the total loss $\mathcal{L}_{\tt total}$ used by OMAN takes the form
\begin{equation}
    \label{eq:loss}
\mathcal{L}_{\tt total} = \mathcal{L}_{\tt OT} + \mathcal{L}_{\tt cls} + \mathcal{L}_{\tt KL}\,.
\end{equation}

\section{Experiments }
\label{sec:exp}

\subsection{Datasets and Metrics}
\label{ssec:exp1}
\textbf{Datasets.} Following previous work~\cite{DRNet}, we choose the SenseCrowd~\cite{SenseCrowd} and CroHD~\cite{HeadHunter-T} datasets. CroHD is an MOT dataset with $9$ videos ($4$ for training, and $5$ for testing), involving crowded indoor and outdoor scenes. SenseCrowd, consisting of $634$ videos, has both sparse and dense scenarios with head coordinates and ID annotations. For VIC, we transform the ID annotations 
into inflow/outflow labels.

\vspace{5pt}
\noindent\textbf{Metrics.} We use the standard VIC metric Weighted Relative Absolute Errors (WRAE) to quantify the performance, defined by
\begin{equation}
    \label{eq:WRAE}
    WRAE=\sum_{i=1}^{K}\frac{T_{i}}{\sum_{j=1}^{K}T_{j}}\frac{|N_{i}-\hat{N}_{i}|}{N_{i}}\times100\%\,,
\end{equation}
where $K$ represents the video number, $T_{i}$ denotes the length of the $i$-th video sequence, $N_{i}$ is the ground-truth pedestrian count of the $i$-th video, and $\hat{N}_{i}$ is the predicted one.
Since VIC also closely relates to counting, generic counting metrics such as Mean Absolute Error (MAE) and Mean Squared Error (MSE) are reported as well. 

\subsection{Implementation Details }
\label{ssec:exp2}

\textbf{Training Details.} For pedestrian localization, we adopt the framework of Point quEry Transformer (PET)~\cite{PET}. For a fair comparison with CGNet~\cite{CGNet}, the ConvNext-S backbone is used. 
The learning rate is set to $1e^{-5}$ for the backbone and $1e^{-4}$ for other network modules. All other training details are kept the same as CGNet. Our model is 
trained on $4$ RTX $3090$ GPUs with the batch size set to $1$.

\vspace{5pt}
\noindent\textbf{Testing Details.} 
Following previous work~\cite{DRNet}, the frame sampling interval $\sigma$ is set to $3$s during testing.

\begin{table}[!t] \footnotesize
    \centering
    \caption{\textbf{Results on SenseCrowd dataset.} D$0$$\sim$D$4$ represents different crowd densities, ranging [$0$,$50$), [$50$,$100$), [$100$,$150$), [$150$,$200$) and [$200$,$+\infty$). Best performance is in \textbf{boldface}, and second best is \underline{underlined}.}
    \label{tab:SENSE}
    \renewcommand{\arraystretch}{1} 
    \addtolength{\tabcolsep}{-4.5pt} 
    \begin{tabular}{@{}lcccccccc@{}}
        \toprule
        \multirow{2}{*}{Method} & \multirow{2}{*}{MAE$\downarrow$} & \multirow{2}{*}{MSE$\downarrow$} & \multirow{2}{*}{WRAE (\%)$\downarrow$} & \multicolumn{5}{c}{Density\textbf{ }level (MAE$\downarrow$)} \\
        \cmidrule{5-9}
        & & & & D0 & D1 & D2 & D3 & D4 \\
        \midrule
        FairMOT~\cite{FairMOT} & 35.4 & 62.3 & 48.9 & 13.5 & 22.4 & 67.9 & 84.4 & 145.8 \\
        HeadHunter-T~\cite{HeadHunter-T} & 30.0 & 60.6 & 44.9 & 8.9 & 17.3 & 56.2 & 96.4 & 131.4 \\
        \midrule
        LOI~\cite{LOI-2016} & 24.7 & 33.1 & 37.4 & 12.5 & 25.4 & 39.9 & 104.1 & 115.2 \\
        \midrule
        DRNet~\cite{DRNet} & 12.3 & 24.7 & 12.7 & \underline{4.1} & 8.0 & 23.3 & 50.0 & 77.0 \\
        CGNet~\cite{CGNet} & \underline{8.9}& 17.7& 12.6 & 5.0 & \textbf{5.8} &\textbf{ 8.5} & 25.0 & 63.4 \\ 
        PDTR~\cite{PDTR} & 9.6 & \underline{17.6} & \underline{11.4} & 4.6 & 6.8 & 14.7 & \underline{23.6} & \underline{60.6} \\ 
        \midrule
        OMAN (Ours) & \textbf{8.3}& \textbf{15.4}& \textbf{11.1}& \textbf{3.9}& \underline{6.2}& \underline{13.2}& \textbf{20.3}& \textbf{48.2}\\
        \bottomrule
    \end{tabular}
\end{table}

\begin{table*}[!t] \small
    \centering
    \caption{\textbf{Results on CroHD dataset.} CroHD$11$$\sim$CroHD$15$ represents five scenes, with $133$, $737$ ,$734$, $1040$, and $321$ as ground-truth counts respectively. Values in `( )' represents the deviation from ground-truth counts, and `--' is due to unreleased code. Best performance is in \textbf{boldface}, and second best is \underline{underlined}.}
    \label{tab:CroHD}
    \renewcommand{\arraystretch}{0.9} 
    \addtolength{\tabcolsep}{1.7 pt}
    \begin{tabular}{@{}lccccccccc@{}}
        \toprule
        \multirow{3}{*}{Method} & \multirow{3}{*}{MAE$\downarrow$} & \multirow{3}{*}{MSE$\downarrow$} & \multirow{3}{*}{WRAE(\%)$\downarrow$} & \multicolumn{5}{c}{MAE$\downarrow$ on five testing scenes} \\
        \cmidrule{5-9}
        & & & & CroHD11 & CroHD12 & CroHD13 & CroHD14 & CroHD15 \\
        & & & & 133 & 737 & 734 & 1040 & 321 \\
        \midrule
        PHDTT~\cite{PHDTT} & 2130.4 & 2808.3 & 401.6 & 380 (247) & 4530 (3793) & 5528 (4794) & 1531 (491) & 1648 (1327) \\
        FairMOT~\cite{FairMOT} & 256.2 & 300.8 & 34.1 & 144 (11) & 1164 (427) & 1018 (284) & 632 (408) & 472 (1000) \\
        HeadHunter-T~\cite{HeadHunter-T} & 253.2 & 351.7 & 32.7 & 198 (65) & 636 (101) & 219 (515) & 458 (582) & \textbf{324 (3)} \\
        ByteTrack~\cite{ByteTrack} & 213.2 & 340.2 & 30.8 & 160 (27) & \textbf{761 (24)} & 1467 (733) & 897 (143) & 460 (139) \\
        \midrule
        LOI~\cite{LOI-2016} & 305.0 & 371.1 & 46.0 & 72.4 (60) & 493.1 (243) & 275.3 (458) & 409.2 (630) & 189.8 (131) \\
        \midrule
        DRNet~\cite{DRNet} & 141.1 & 192.3 & 27.4 & 164.6 (31) & 1075.5 (338) & 752.8 (18) & 784.5 (255) & 382.3 (61) \\
        CGNet~\cite{CGNet} & 75.0 & 95.1 & 14.5 &\underline{-- (7)} & -- (72) & \underline{-- (14)} & -- (144) & -- (138) \\
        FMDC~\cite{FMDC} & \underline{54.2} & \underline{61.7} & \underline{10.7} & \textbf{138.9 (5)} & 664.3 (73) & 818.0 (84) & \textbf{1005.8 (35)} & 394.9 (73) \\ 
        PDTR~\cite{PDTR} & 60.6 & 73.7 & 12.7 & 109 (24) & \underline{678 (59)} & \textbf{729 (5)} & 935 (105) & 431 (110) \\ 
        \midrule
        OMAN (Ours) & \textbf{38.6} & \textbf{50.5} & \textbf{8.3 }& 142 (9) & 636 (101) & 715 (19) & \underline{1003 (37)} & \underline{348 (27)} \\ 
        \bottomrule
    \end{tabular}
\end{table*}

\subsection{Experimental Results }
\label{ssec:exp3}
\begin{figure}[!t]
  \centering
  \centerline{\includegraphics[width=8.5cm]{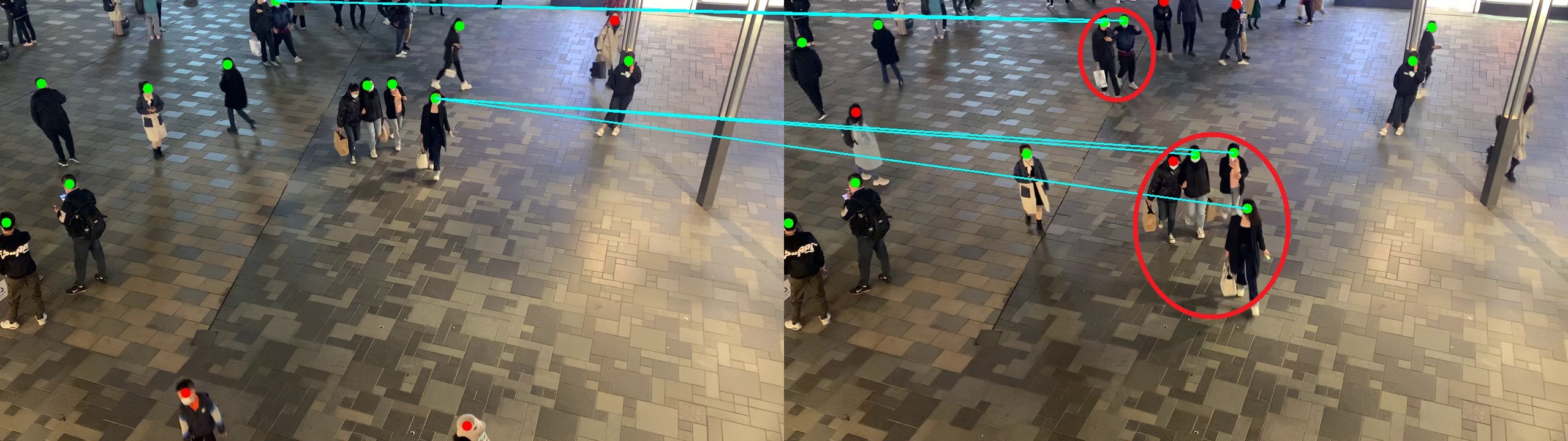}}\vspace{-10pt}
  \caption{\textbf{Visualization of O2M matching.} Red dots indicate inflows/outflows, green dots indicate shared pedestrians, red circles represent groups, and blue lines indicate matches.}
    \label{fig:vis}
\end{figure}

\textbf{Results on SenseCrowd.} Results are reported in Table~\ref{tab:SENSE}. It shows that 
OMAN surpasses the 
VIC baselines in general in 
all metrics, 
achieving 
$8.3$ MAE, $15.4$ MSE, and $11.1$\% WRAE. Specifically, our model 
reports the best results on D$0$, D$3$ and D$4$, including both the sparsest and the densest scenes, and is also with the second-best results achieved on D$1$ and D$2$. 
In addition, Fig.~\ref{fig:vis} visualizes that OMAN indeed implements the O2M matching. 

\vspace{5pt}
\noindent\textbf{Results on CroHD.} Experimental results on the CroHD dataset is shown in Table \ref{tab:CroHD}. We observe that our proposed OMAN is also the best performing one. Generally, by our method, MAE, MSE and WRAE are reduced by $28.8$\%, $18.2$\%, and $22.4$\%, respectively. CroHD$11$$\sim$CroHD$15$ represent five different scenes with different densities, while the differences between predicted and true values in different scenes are generally small, indicating that our model is robust to various densities of the pedestrian.

\subsection{Ablation Study}
\label{ssec:exp4}
\begin{table}[!t]
    \caption{\textbf{Ablation study on SenseCrowd dataset}.}
    \centering
    \label{tab:ablation-methods}
    \renewcommand{\arraystretch}{0.8} 
    \addtolength{\tabcolsep}{3.5 pt} 
    \begin{tabular}{@{}cccccc@{}}
        \toprule
        ICG&OMPM&$\mathcal{L}_{\tt KL}$&MAE$\downarrow$&MSE$\downarrow$&WRAE$\downarrow$ \\
        \midrule
        &&&16.88&27.18&22.00\% \\
        \Checkmark&&&9.62&19.79&12.89\% \\
        &\Checkmark&&8.75&16.43&11.80\% \\
        \Checkmark&\Checkmark&&8.34&15.97&11.41\% \\
        \Checkmark&\Checkmark&\Checkmark&8.26&15.42&11.07\% \\
        \bottomrule
    \end{tabular}
\end{table}

\textbf{Impact of Proposed Modules.} In Table~\ref{tab:ablation-methods}, we 
ablate our approach to justify the contribution of the modules. One can observe that, the implicit group contexts provided by ICG and the O2M strategy contribute to the most improvements.

\begin{figure}[!t]
  \centering
  \centerline{\includegraphics[width = 7cm]{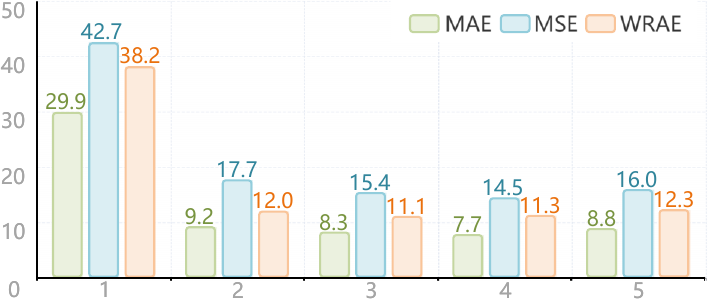}}
  \vspace{-10pt}
  \caption{\textbf{Performance of OMPM with different layer numbers.} The horizontal axis indicates the number of layers.}
    \label{fig:mlp}
\end{figure}

\vspace{5pt}
\noindent\textbf{Performance of Different OMPM Modules.} Since we simply employ an MLP to implement the O2M matching,
we conduct an experiment on the SenseCrowd dataset to determine the optimal layer configuration. As shown in Fig.~\ref{fig:mlp}, the WRAE reaches its minimum when the number of MLP layers is equal to three. While the MAE and MSE metrics are slightly worse, we still report the three-layer results considering that the main metric of VIC is WRAE.

\section{Conclusion}
\label{sec:conc}
In this work, we introduce a simple but effective model called OMAN for VIC. Unlike previous O2O-based approaches, we first generate token-to-token similarity context and execute implicit matching with an ICG module implemented mainly based on standard self-attention. Then we leverage an OMPM module implemented by simple MLP to implement O2M assignment. We conduct experiments on two standard benchmark datasets and report the state-of-the-art performance on both. Due to the simplicity of implementation and a fresh perspective, we believe OMAN will serve as a strong baseline for VIC with ample scope for improvement.

\vspace{5pt}
\noindent\textbf{Acknowledgement.}
This work is supported by Hubei Provincial Natural Science Foundation of China under Grant No. 2024AFB566.

\vfill\pagebreak
\label{sec:refs}

\bibliographystyle{IEEEbib}
\bibliography{strings,refs}

\end{document}